\documentclass[lettersize,journal]{IEEEtran}
\usepackage{amsmath,amsfonts}
\usepackage{algorithmic}
\usepackage{array}
\usepackage[caption=false,font=normalsize,labelfont=sf,textfont=sf]{subfig}
\usepackage{textcomp}
\usepackage{stfloats}
\usepackage{url}
\usepackage{verbatim}
\usepackage{graphicx}
\usepackage{cite}
\usepackage{booktabs}
\usepackage{multirow}
\usepackage{orcidlink}
\usepackage{amssymb}
\usepackage{makecell}
\usepackage{float}
\usepackage{CJKutf8}
\usepackage[most]{tcolorbox}

\tcbset{
  enhanced,
  colback=gray!3!white,
  colframe=black!40,
  coltitle=white,
  fonttitle=\bfseries,
  colbacktitle=gray!60!black,
  boxrule=0.4pt,
  arc=2pt,
  left=3pt,right=3pt,top=4pt,bottom=4pt,
  listing only,
  listing options={
    basicstyle=\ttfamily\footnotesize,
    breaklines=true,
    columns=fullflexible,
    numbers=none
  }
}

\usepackage{pifont}
\usepackage[ruled,linesnumbered]{algorithm2e}
\usepackage{xcolor}
\definecolor{mycolor}{RGB}{234, 107, 102}
\definecolor{mycolor1}{RGB}{185, 224, 165}
\definecolor{mycolor2}{RGB}{255, 255, 153}

\begin{document}

\title{EVE: Towards End-to-End Video Subtitle Extraction with\\Vision-Language Models}

\author{Haiyang Yu, Mengyang Zhao,  Jinghui Lu, Ke Niu, Yanjie Wang, Weijie Yin, Weitao Jia, Teng Fu, Yang Liu \\ Jun Liu, Hong Chen~\IEEEmembership{Fellow,~IEEE}
\thanks{
Haiyang Yu, Mengyang Zhao, and Jinghui Lu are co-first authors.
Corresponding authors:  Ke Niu; Yang Liu.

Haiyang Yu, Mengyang Zhao, Ke Niu, and Teng Fu are with the College of Computer Science and Artificial Intelligence, Fudan University, Shanghai 200433, China  (e-mail: hyyu20@fudan.edu.cn, myzhao20@fudan.edu.cn, kniu22@m.fudan.edu.cn, tfu23@m.fudan.edu.cn).

Jinghui Lu, Yanjie Wang, Weijie Yin, Weitao Jia are with ByteDance Inc., Yangpu District, Shanghai 200082, China (e-mail: lujinghui@bytedance.com, wangyanjie.prince@bytedance.com, yinweijie@bytedance.com,  jiaweitao@bytedance.com, liyang.712@bytedance.com).

Yang Liu is with the College of Electronic and Information Engineering, Tongji University, Shanghai 201804, China  (e-mail: yang\_liu@ieee.org).

Jun Liu is with School of Computing and Communications, Lancaster University, Lancaster, LA1 4WA, UK (e-mail: j.liu81@lancaster.ac.uk).

Hong Chen is with College of Electronic and Information Engineering, Tongji University, Shanghai 201804, China (chenhong2019@tongji.edu.cn).

}
}

\markboth{IEEE TRANSACTIONS ON IMAGE PROCESSING,November~2025}%
{Shell \MakeLowercase{\textit{et al.}}: A Sample Article Using IEEEtran.cls for IEEE Journals}


\maketitle

\begin{abstract}
Video subtitles play a crucial role in short videos and movies, as they not only help models better understand video content but also support applications such as video translation and content retrieval. 
Existing video subtitle extraction methods typically rely on multi-stage frameworks, where errors accumulate across stages and temporal dependencies are underutilized due to frame-wise processing. 
Moreover, although some Large Vision-Language Models (LVLMs) possess strong OCR capabilities, predicting accurate timestamps for subtitle texts remains challenging. 
To this end, we propose an \underline{E}nd-to-end \underline{V}ideo subtitle \underline{E}xtraction framework based on LVLMs, named \textbf{EVE}, which can output subtitles and their timestamps simultaneously.  Specifically, we introduce a dual-branch \underline{S}patiotemporal \underline{S}ubtitle-\underline{S}alient  (S\textsuperscript{3}) Module that serves as an adapter for LVLMs, capable of representing subtitle-related content and considering inter-frame correlations using only a small number of tokens.  Within this module, the \textbf{Spatial Semantic Context Aggregate} branch aggregates high-level global semantics to provide spatial visual contextual information, while the \textbf{Temporal Subtitle Token Query} branch explicitly queries subtitle-relevant tokens while considering temporal correlation across frames. The small number of tokens retained by the S\textsuperscript{3} module are fed to the language model, which then directly outputs the subtitle text along with its timestamps. Furthermore, we construct the first large-scale dataset dedicated to video subtitle extraction, \textbf{ViSa}, containing over 2.5M videos with  timestamped and bilingual annotation, thereby providing the community with a well-organized training and evaluation benchmark.
Extensive experiments on ViSa demonstrate that EVE significantly outperforms existing open-source tools and LVLMs in both subtitle recognition accuracy and temporal alignment, achieving state-of-the-art performance.

\end{abstract}

\begin{IEEEkeywords}
Video Subtitle Extraction, Large Vision-Language Models, Token Compression
\end{IEEEkeywords}

\section{Introduction}

With the rapid growth of video as the dominant medium for information dissemination, video understanding has become a central topic in multi-modal research~\cite{wu2021towards,wu2024deep,tip-video-sum1, tip-video-sum2}. 
Among the various modalities embedded in videos, textual information—particularly \textit{subtitles}—serves as an indispensable bridge between visual and linguistic domains, providing explicit and contextually aligned semantic information. 
Accurate subtitle extraction not only enhances the ability of Vision-Language Models (VLMs) to comprehend video content but also enables a wide range of downstream applications, such as cross-lingual subtitle translation and video content retrieval \cite{tip-video-ret}. 
Therefore, video subtitle extraction is of great importance both as a fundamental research problem and as a key enabling technology for practical video understanding applications.

\begin{figure}[t]
  \centering
   \includegraphics[width=0.98\linewidth]{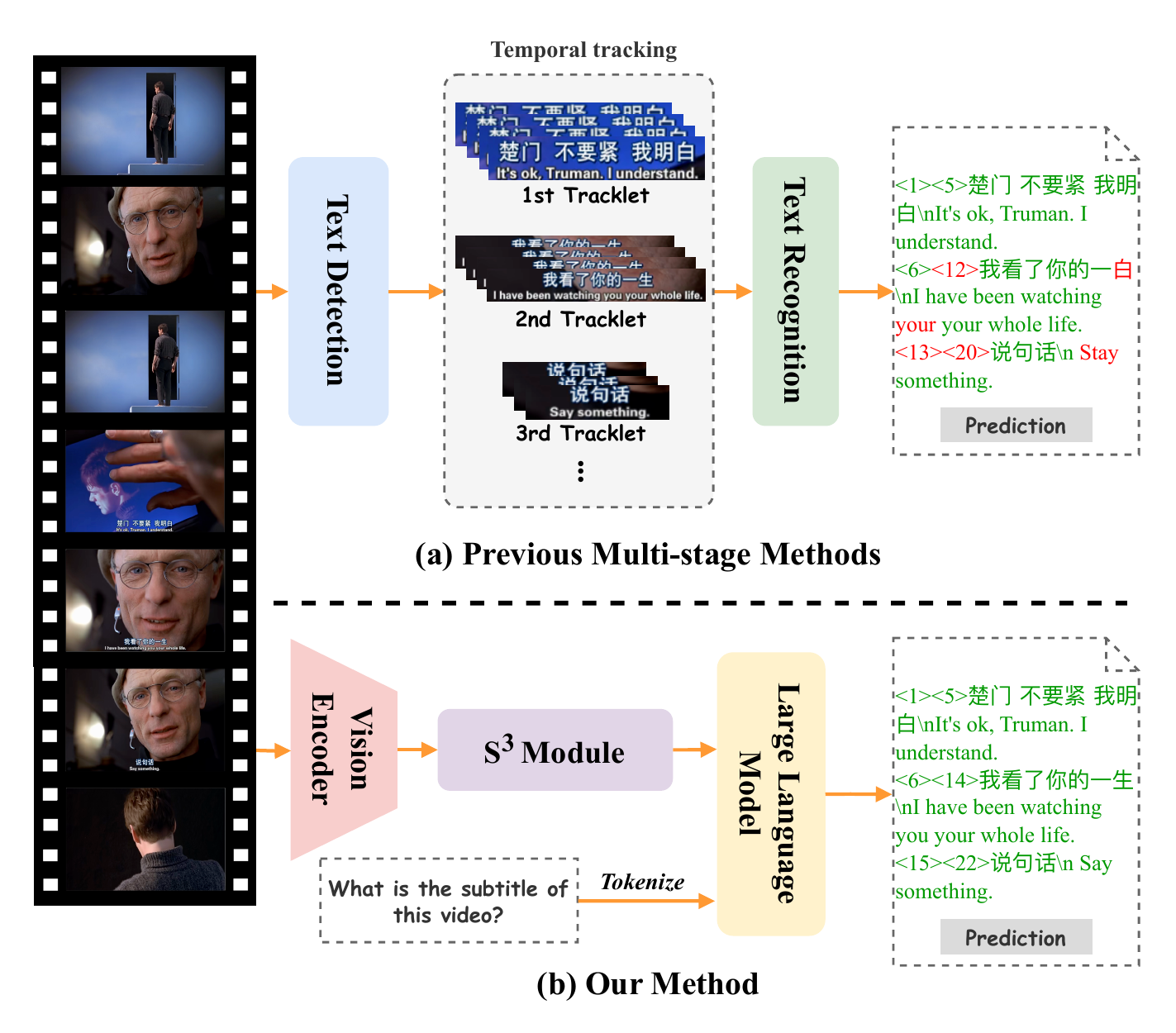}

   \caption{Comparison of the proposed method with previous methods on the pipeline. Most previous methods (a) rely on a multi-stage framework, where prediction error accumulates at each stage, influencing the final prediction. In contrast, the proposed EVE (b) is an end-to-end video subtitle extraction method.}
   \label{fig:intro}
\end{figure}

Previous video subtitle extraction methods can be broadly categorized into two groups: OCR-based approach and speech-based approach. The OCR-based approach~\cite{elshahaby2022end,yan2020end} often involves multiple stages, such as text detection, temporal tracking, and text recognition for subtitle regions as shown in Figure~\ref{fig:intro} (a). Although these methods achieve reasonable performance, the multi-stage approach is susceptible to error accumulation. For example, if the output of text detection is inaccurate, text recognition models can hardly recognize the corresponding texts correctly. In addition, they process each frame independently without considering the temporal information between frames. Different from the OCR-based approach, the speech-based approach~\cite{papi2023direct,mathur2015generating} directly transcribes the speech of videos to texts and timestamps the texts to generate subtitles. However, these methods are sensitive to noise in audios and often struggle to align the recognized subtitles with visual text cues in videos.

Recent advances in Vision-Language Models (VLMs) have substantially improved the performance of various video-related tasks~\cite{kim2024image,wang2024videotree,xu2024pllava,wu2023bidirectional}. 
However, the task of video subtitle extraction remains underexplored. To our knowledge, there is no efficient end-to-end VLM-based frameworks or comprehensive datasets available.  In addition, video subtitle extraction poses two unique challenges that distinguish it from general video understanding. First, subtitles appear in very small spatial regions and depend on fine-grained visual cues, which are easily lost under the token compression and global pooling strategies typical of existing LVLMs. Second, subtitles persist across multiple frames with gradual transitions, requiring high-fidelity modeling of both their textual content and temporal boundaries. Such temporal coherence cannot be captured by frame-wise OCR or captioning, often resulting in inconsistent or misaligned subtitles.

To fill this gap, we propose an \underline{E}nd-to-end \underline{V}ideo subtitle \underline{E}xtraction framework, termed EVE. EVE comprises three major components: a vision encoder, a \underline{S}patiotemporal \underline{S}ubtitle-\underline{S}alient  (S\textsuperscript{3}) module, and a large language model (LLM). 
The S\textsuperscript{3} module consists of two branch and acts as an adapter that refines frame-level visual features from two complementary perspectives while retaining only a small number of tokens. 
Specifically, the Spatial Semantic Context Aggregate (SSCA) branch aggregate high-level global semantics to provide spatial contextual cues for subtitle extraction. 
Meanwhile, the Temporal Subtitle Token Query (TSTQ) branch explicitly queries subtitle-relevant tokens inspired by the Q-former \cite{li2023blip}. Q-former is designed for general vision-language alignment,
while TSTQ explicitly targets subtitle-bearing regions and enforces temporal consistency. 
The outputs from both branches are individually projected into the language model’s embedding space and then concatenated before being fed into the LLM to generate timestamped subtitle texts. To support bilingual  (\textit{i.e.}, Chinese and English)  video subtitle extraction, we employ Qwen2~\cite{yang2024qwen2} as the large language model, leveraging its strong cross-lingual understanding and generation capabilities to ensure subtitle accuracy.

Although several datasets \cite{elshahaby2022end,1227749,6628859,Wang20111457} have been used in existing video subtitle extraction method, they suffer from issues such as limited scale, lack of timestamp annotations, or derivation from images rather than videos. This renders these datasets vastly inferior to those for other relevant tasks, rendering them inadequate to meet the current data requirements of LVLM.
Therefore, we construct a large-scale benchmark dataset, ViSa, containing over 2.5M videos collected from CNVid~\cite{gan2023cnvid} and MovieChat~\cite{song2024moviechat} for pre-training, supervised fine-tuning, and testing. 
ViSa provides bilingual annotations of subtitle texts and timestamps, offering a valuable resource for systematic evaluation and future research on video subtitle extraction. 

Extensive experiments on ViSa demonstrate that EVE significantly outperforms existing open-source tools and LVLM baselines in both subtitle recognition accuracy and temporal alignment, setting a new state-of-the-art for video subtitle extraction. In summary, the main contributions of this work are as follows:

\begin{itemize}
    \item We propose an \underline{E}nd-to-end \underline{V}ideo subtitle \underline{E}xtraction framework, EVE, built upon LVLMs to directly generate timestamped subtitles from video input.
    
    \item We design a \underline{S}patiotemporal \underline{S}ubtitle-\underline{S}alient  (S\textsuperscript{3}) Module, which refines visual features through spatial semantic aggregation and temporal subtitle token querying to effectively capture subtitle-salient representations and consider temporal consistency with only a small number of tokens.

    \item We construct the first large-scale dataset, ViSa, containing over 2.5M videos with bilingual subtitle text and timestamp annotations, providing a valuable resource for systematic evaluation and future research in this field.
    
    \item Extensive experiments on ViSa demonstrate that EVE significantly outperforms existing open-source tools and LVLM baselines in video subtitle extraction task.
\end{itemize}
The remainder of this paper is organized as follows:
Section~\ref{sec:related} reviews existing  work on video subtitle extraction.
Section~\ref{sec:Methodology} presents the overall architecture of EVE, including the vision encoder, the proposed S\textsuperscript{3} module, and the training and inference procedures.
Section~\ref{sec:Dataset} outlines the characteristics of the ViSa dataset and its construction process.
Section~\ref{exp} reports extensive experiments, including comparisons with existing tools and LVLMs, ablation studies, visualizations, and hyperparameter analyses.
Finally, Section~\ref{Conclusion} concludes the paper, discusses the limitations of the proposed method, and outlines potential future directions.

\section{Related Work}
\label{sec:related}
\subsection{Video Subtitle Extraction}

Subtitles, as supplementary information in videos, play a crucial role in video understanding. 
Existing approaches to video subtitle extraction can be broadly categorized into two groups: OCR-based and speech-based approaches. 

\subsubsection{OCR-based approach} OCR-based approaches typically treat subtitle extraction as a multi-stage pipeline. 
For example, Elshahaby \textit{et al.} ~\cite{elshahaby2022end} proposed an system that first performs text localization to identify regions of interest within video frames, followed by text detection and recognition on the localized areas. 
Similarly, Yan \textit{et al.}~\cite{yan2020end} adopted a comparable framework, employing the Connectionist Text Proposal Network (CTPN)~\cite{tian2016detecting} for subtitle region detection. 
However, such methods process each frame independently, making it difficult to leverage the temporal and contextual information across frames. 
Moreover, the sequential nature of detection and recognition stages often leads to error accumulation, where inaccurate detection can significantly degrade the final recognition performance. VideOCR\footnote{https://github.com/devmaxxing/videocr-PaddleOCR} is an open-source video subtitle extraction tool developed based on the OCR tool PaddleOCR \footnote{https://github.com/PaddlePaddle/PaddleOCR}.

\subsubsection{Speech-based approach}

Mathur \textit{et al.} \cite{mathur2015generating} presents an offline pipeline that extracts audio from video, applies speech recognition, and converts the recognized speech into time-aligned subtitle files.
Ramani \textit{et al.} ~\cite{ramani2020automatic} extract subtitles by transcribing the audio stream of videos using speech recognition models and then associating the transcribed text with timestamps. Sara \textit{et al.} \cite{papi2023direct} proposes an end-to-end direct speech translation model that generates time-aligned target-language subtitles from source audio.
While effective for clear speech, these methods directly treat transcribed speech as subtitles, which can cause mismatches with the actual on-screen text—particularly in short-form videos (\textit{e.g.}, YouTube Shorts, TikTok) where subtitle alignment with speech is often imperfect. 

In addition to the aforementioned two types of methods, there are also tools available for extracting video subtitles. PyVideoTran \footnote{https://github.com/jianchang512/pyvideotrans} is a fully automatic video/audio translation tool that transcribes speech, generates and translates subtitles, which also supports batch processing of video or audio files into precisely time-coded SRT subtitle files.  Commercial cloud services (\textit{e.g.}, OpenVision\footnote{https://vision.aliyun.com/}, GhostCut\footnote{https://cn.jollytoday.com/home/}) have also been developed for video subtitle extraction.

\subsection{Large Vision-Language Models}

In recent years, Large Vision-Language Models (LVLMs)~\cite{wang2024qwen2,yao2024minicpm,zhang2023video,li2025llama,huang2025lita} have demonstrated remarkable capabilities in instruction following and multimodal reasoning, enabling them to capture on-screen text or subtitles when prompted with appropriate instructions.
Unlike conventional OCR-based methods, these models can leverage global visual-semantic context and learn high-level associations between video content and textual cues.
However, their subtitle extraction performance remains limited in both accuracy and temporal alignment, mainly due to the lack of task-specific training for fine-grained subtitle localization.
Most existing LVLMs are optimized for holistic video captioning or question answering \cite{song2024moviechat,wang2024qwen2} rather than precise frame-level text prediction.
Consequently, they tend to produce scene-level descriptions or paraphrased summaries instead of exact subtitle transcripts.
This highlights the need for a task-specific framework that can jointly model spatial and temporal semantics to achieve accurate and temporally consistent subtitle extraction.

\begin{figure*}[t]
  \centering
   \includegraphics[width=1.0\linewidth]{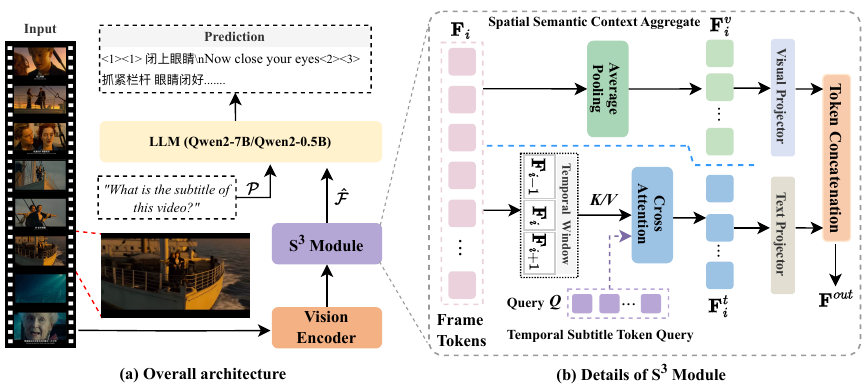}

   \caption{Overall architecture of EVE (a) and details of S\textsuperscript{3} module (b). EVE contains three modules: a vision encoder, an adapter and a large language model. We propose a novel adapter called S\textsuperscript{3} module, consisting of a spatial semantic context aggregating and a temporal subtitle token querying branches. To enhance the correlations between adjacent frames, we  also introduce a temporal window mechanism.}
   \label{fig:overall}
\end{figure*}

\subsection{Video Token Compression}

The proposed S\textsuperscript{3} module aims to extract subtitle-relevant representations from dense visual tokens and retain only a compact set of key tokens for the large language model. 
This mechanism is conceptually related to the token compression strategies widely adopted in recent LVLMs, which also focus on preserving the most informative tokens while reducing redundancy to enhance efficiency. Existing token compression methods can be roughly divided into two categories. 
The first category relies on  token merging, where similar visual tokens are grouped or averaged to reduce redundancy, as exemplified by ToMe~\cite{bolyatoken} and its variants~\cite{liang2022not, liusimple}. 
The second category employs query-based token selection, which introduces a small set of learnable queries to extract compact visual representations through cross-attention, similar to the Q-Former module~\cite{li2023blip}. 
This approach has been adopted by several video-related VLMs such as LLaMA-VID~\cite{li2025llama}, TimeChat~\cite{ren2024timechat}, and Video-LLaMA~\cite{zhang2023video}. 
However, studies such as DeCo~\cite{yao2024deco} have shown that Q-Former-based compression often requires large-scale training data and can suffer from unstable convergence. 
In contrast, simple operations like average pooling have demonstrated surprisingly strong performance in some cases, as reported in LITA~\cite{huang2025lita}, Video-ChatGPT~\cite{maaz2023video}, and Oryx ~\cite{liu2024oryx}. 
Although our S\textsuperscript{3} module shares the same philosophy of extracting a minimal yet informative token set, it is specifically designed for the video subtitle extraction task. Instead of general-purpose visual redundancy reduction, S\textsuperscript{3} focuses on identifying tokens that are semantically and temporally relevant to subtitles.

\section{Methodology}
\label{sec:Methodology}

\subsection{Architecture}
The overall architecture of the proposed \underline{E}nd-to-end \underline{V}ideo subtitle \underline{E}xtraction framework (EVE) is illustrated in Figure~\ref{fig:overall}. 
EVE consists of three main components: a vision encoder, a \textbf{\underline{S}}patiotemporal \textbf{\underline{S}}ubtitle-\textbf{\underline{S}}alient (S\textsuperscript{3}) Module, and a large language model (LLM). 
Given an input video $\mathcal{V}=\{\mathbf{I}_0, ..., \mathbf{I}_{N-1}\}$, each frame $\mathbf{I}_i$ ($i \in \{0, 1, ..., N-1\}$) is first processed by the vision encoder to extract frame-level visual features $\mathbf{F}_i$. 
The resulting feature sequence can be denoted as $\mathcal{F}=\{\mathbf{F}_0, ..., \mathbf{F}_{N-1}\}$, where $\mathbf{F}_i \in \mathbb{R}^{(H\times W)\times C}$ represents the dense visual tokens of the $i$-th frame. To obtain subtitle-relevant representations, we propose the S\textsuperscript{3} module, which comprises two components: the Spatial Semantic Context Aggregation (SSCA) and the Temporal Subtitle Token Query (TSTQ). The former aggregates global semantics from spatial contexts, while the latter queries subtitle-relevant tokens through temporal salience modeling.  The output of S\textsuperscript{3} is projected into the language model's embedding space and denoted as $\mathbf{F}^{out}$ after concatenation.
This representation, conditioned on an instruction prompt $\mathcal{P}$, is subsequently fed into the large language model to produce the subtitle sequence $\mathcal{S}={S_0, \ldots, S_{T-1}}$, where each subtitle $S_i$ is formatted as ``$\langle b_i\rangle \langle e_i\rangle s_i$'', with $b_i$ and $e_i$ denoting its start and end frames.

In the following subsections, we describe the vision encoder, the proposed S\textsuperscript{3} Module, and the large language model in detail.

\subsection{Vision Encoder} In recent years, various Transformer-based vision backbones are proposed, including ViT~\cite{dosovitskiy2020image}, Swin-Transformer~\cite{liu2021swin}, EViT~\cite{liang2022not}, Dynamic ViT~\cite{rao2021dynamicvit}, etc. Among these Transformer-based backbones, we select Swin-Transformer pretrained in Donut~\cite{kim2021donut} as the vision encoder of EVE for two reasons: (1) \textit{Better initialization for OCR}. Since the Swin-Transformer backbone of Donut has been pre-trained on numerous document datasets for multiple OCR tasks, it can provide better initialization to enhancing the text perception ability of EVE. In addition, based on the pre-training Swin-Transformer of Donut, our model can converge faster during the pre-training phase. (2) \textit{Higher time efficiency}. Benefiting from the introduced shifted window-based self-attention module in Swin-Transformer, it shows better time efficiency, which can accelerate extracting visual features of frames in videos. 

\subsection{Spatiotemporal Subtitle-Salient Module}
The proposed S\textsuperscript{3} module comprises the following two branches:
\subsubsection{Spatial Semantic Context Aggregate}
Subtitle generation is not a pure text-recognition problem but is inherently guided by the visual semantics of each frame. Movie and short-video subtitles often correspond to high-level scene cues—such as character interactions, camera composition, and background layout—that provide strong prior information about whether subtitles appear and how they should be phrased. To capture such global semantics in an efficient and robust manner, we introduce the Spatial Semantic Context Aggregate (SSCA) branch.

SSCA summarizes the frame feature $\mathbf{F}_i$ into a compact layout-aware representation $\mathbf{F}^v_i \in \mathbb{R}^{p\times p\times C}$ using spatial average pooling:
\begin{equation}
    \mathbf{F}^v_i = \texttt{AvgPool}(\mathbf{F}_i).
\end{equation}
Although the operation is simple, it serves a clear and task-aligned role: it suppresses low-level background noise while preserving high-level semantic structure—such as scene geometry, speaker regions, and global motion cues—that influence subtitle presence and content. SSCA thus provides a stable global prior that complements the text-focused modeling of the TSTQ branch.

\subsubsection{Temporal Subtitle Token Query}
While SSCA captures holistic scene cues, accurate subtitle extraction further requires identifying fine-grained visual evidence related to on-screen text. To this end, we adopt a lightweight query-based design inspired by Q-Former~\cite{li2023blip}. A set of learnable query embeddings $\mathbf{Q}$ are used as latent probes that attend to subtitle-salient regions in the dense feature map. Each query specializes in capturing complementary cues—such as text contours, stroke patterns, or high-contrast regions—by performing cross-attention with $\mathbf{F}_i$:
\begin{equation}
    \mathbf{F}^t_i = \texttt{CrossAttention}(\mathbf{Q}, \mathbf{F}_i).
\end{equation}

Unlike general vision–language alignment, subtitle extraction requires strong temporal stability because subtitles usually persist across several frames and gradually fade in or out. To enhance robustness, we extend the attention context to a temporal window consisting of the current frame and its neighbors:
\begin{equation}
    \mathbf{F}^t_i = \texttt{CrossAttention}(\mathbf{Q}, [\mathbf{F}_{i-1},\mathbf{F}_i,\mathbf{F}_{i+1}]).
\end{equation}
This temporal aggregation enables TSTQ to track slowly transitioning subtitle patterns, reduce frame-wise inconsistencies, and generate subtitle-related tokens that are both temporally consistent and visually discriminative.

\subsubsection{Feature Alignment and Fusion}
The SSCA and TSTQ branches provide two complementary perspectives on subtitle perception: SSCA encodes global spatial semantics, while TSTQ extracts localized, temporally coherent subtitle tokens. Although both originate from the same visual encoder, their feature distributions vary substantially—one emphasizes holistic context, and the other emphasizes fine-grained text evidence.

To align these heterogeneous representations with LLM, we introduce the separate projector (SP) strategy :
\begin{equation}
    \hat{\mathbf{F}}^v_i = \texttt{Proj}_v(\mathbf{F}^v_i), \quad
    \hat{\mathbf{F}}^t_i = \texttt{Proj}_t(\mathbf{F}^t_i).
\end{equation}
The separate projector strategy prevents feature interference and preserves the modality-specific characteristics essential for subtitle reasoning.

Finally, the projected global semantics and subtitle-salient tokens across all frames are interleaved to form the unified output sequence:
\begin{equation}
    \mathbf{F}^{out} = [\hat{\mathbf{F}}^v_0, \hat{\mathbf{F}}^t_0, ..., \hat{\mathbf{F}}^v_{N-1}, \hat{\mathbf{F}}^t_{N-1}].
\end{equation}
This fused representation integrates both spatial and temporal cues in a subtitle-aware manner, enabling the language model to produce temporally aligned and semantically faithful subtitles.

\begin{algorithm}[t]
\LinesNumberedHidden
\SetKwInOut{Input}{Input}
\SetKwInOut{Output}{Output}
\Input{Predicted subtitles $\mathcal{S}$=\{$S_0$, $S_1$, ..., $S_{T-1}$\}; OCR results of all frame in the input video $\mathcal{O}$=\{$O_0$, $O_1$, ...., $O_{L-1}$\}, where $O_l$ = \{$\text{o}_0$, $\text{o}_1$,..., $\text{o}_{K_l-1}$\}; Similarity threshold $\mathbf{sim}$; Raw FPS of the input video $\mathbf{f}$; Subtile matching range $\mathbf{r}$.}
\Output{Subtitles in SRT format $\mathcal{S}^\prime$=\{$S^\prime_0$, $S^\prime_1$, ..., $S^\prime_{T-1}$\}, $S^\prime_t$=[$b^\prime_t$, $e^\prime_t$, $s^\prime_t$]}
\BlankLine

\ForEach{subtitle $S$ \textbf{in} $\mathcal{S}$}{
        \textcolor{gray}{\# calculate the coarse timestamp of S } \textcolor{white}{\;}
        $b_c, e_c$ = $\frac{\mathbf{f} \times b}{\text{Sampling Rate}}$, $\frac{\mathbf{f} \times e}{\text{Sampling Rate}}$\textcolor{white}{\;}
        \textcolor{gray}{\# perform matching in the range $\mathbf{r}$ to locate the accurate start timestamp} \textcolor{white}{\;}
        \For{$i$=$b_c-\mathbf{r}$ \KwTo $b_c+\mathbf{r}$}{
            $min_\text{ed}$ = $\infty$\textcolor{white}{\;}
            \ForEach{Text o \textbf{in} $O_i$}{
                \If{\texttt{EditDistance}$(S, o)$ $<$ min$_\text{ed}$}{
                    $min_\text{ed}$ = \texttt{EditDistance}$(S, o)$\textcolor{white}{\;}
                }
            }
            \textcolor{gray}{\# locate the accurate start timestamp $b^\prime$} \textcolor{white}{\;}
            \If{$min_\text{ed} < \mathbf{sim}$}{
                $b^\prime$ = i \textcolor{white}{\;}
                \textbf{break} \textcolor{white}{\;}
            }
    }
    \textcolor{gray}{\# perform the same operation as before to locate the accurate end timestamp $e^\prime$} \textcolor{white}{\;}
    \textcolor{gray}{\# texts in predicted subtitles and converted subtitles are consistent} \textcolor{white}{\;}
    $S^\prime = S$
}

\Return Subtitles in SRT format $\mathcal{S}^\prime$ \textcolor{white}{\;}
\caption{Converting EVE predictions into subtitles with SRT format.}
\label{ocr-matching}
\end{algorithm}

\subsection{Large Language Model} In the past few years, numerous foundation large language models are open-sourced, such as Qwen2~\cite{yang2024qwen2} and LLaMa~\cite{touvron2023llama}. Since the proposed EVE aims at solving bilingual (\textit{i.e.}, English and Chinese) video subtitle extraction, we select Qwen2 as the large language model of EVE. In addition, taking into account the time efficiency in practical applications, we build EVE-7B and EVE-0.5B based on Qwen2-7B and Qwen2-0.5B, respectively. Detail performance comparison is shown in Section~\ref{exp}.

\subsection{Training}

\begin{figure}[t]
  \centering
   \includegraphics[width=1.0\linewidth]{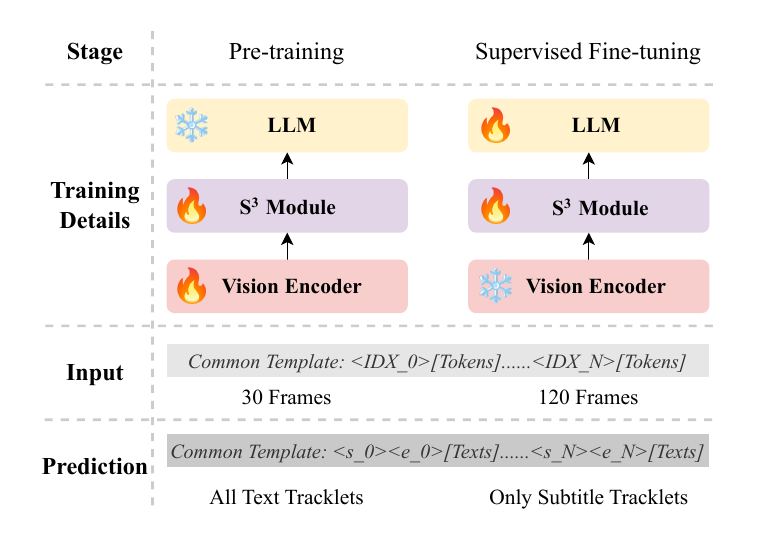}

   \caption{Training details of EVE. The LLM is frozen during pre-training while the vision encoder is frozen during supervised fine-tuning.}
   \label{fig:training}
\end{figure}

The training process of the proposed EVE consists of two stages: a \textit{pre-training} stage and a \textit{supervised fine-tuning} stage. 
We describe these two stages in detail from the following three aspects:

\begin{itemize}
    \item \textbf{Training Details.} 
    The pre-training stage aims to enhance the text perception capability of EVE by enabling it to predict all text tracklets appearing in video frames. 
    As shown in Figure~\ref{fig:training}, during this stage, the large language model is frozen, while the vision encoder and the proposed S\textsuperscript{3} module are jointly optimized. 
    After pre-training, the vision encoder becomes proficient at perceiving textual regions. 
    Consequently, in the supervised fine-tuning stage, the vision encoder is frozen, and only the S\textsuperscript{3} module together with the large language model are trained to distinguish whether the detected texts correspond to subtitles.

    \item \textbf{Input.} 
    Both stages share the same input template. 
    To improve temporal perception, we prepend the corresponding frame index to each frame’s visual tokens. 
    The number of input frames is limited to 30 during pre-training and 120 during supervised fine-tuning.

    \item \textbf{Prediction.} 
    The prediction template is also identical across the two stages, containing the start frame index, end frame index, and  texts. 
    The difference lies in their objectives: the pre-training stage learns to recognize all text tracklets in videos, whereas the supervised fine-tuning stage focuses solely on those corresponding to subtitles.
\end{itemize}

\subsection{Inference}

\subsubsection{Output Format}
We use the SRT files as the model's output, as they are one of the most popular subtitle file formats for video content. An SRT file stores sequential subtitle texts with corresponding timestamps but no audio or video data, resulting in a compact file size. In addition, the SRT file needs to specify the line number, subtitle content, when the text should appear on the video, and when it should disappear to make room for the next line, as shown in Figure~\ref{fig:srt}.

\subsubsection{Frame Sampling and Post-processing} Due to the limitation of EVE in input token length, inference cannot be completed at high frame rates. Therefore, the proposed EVE samples the video at 2 frames per second (FPS) for model input. Although the predicted frame index can directly correspond to a precise timestamp (\textit{i.e.}, hour, minute, and second), it does not represent the precise time at which each subtitle appears and disappears in the video. To output more accurate subtitle content, such as in SRT format, we propose a post-processing method based on OCR results at full frame rate. The algorithm of post-processing is described in Algorithm~\ref{ocr-matching}. Firstly, we use an OCR engine to detect and recognize text in images at full frame rate. Then, we map the frame indices identified by EVE to rough timestamps at full frame rate. Finally, by calculating the edit distance between predicted texts and those within adjacent frames, the precise subtitle timestamp can be obtained.

\begin{figure}[t]
\centering
\begin{CJK*}{UTF8}{gbsn}
\begin{tcolorbox}[
    width=0.8\linewidth,
    title={Example of a bilingual SRT subtitle file},
    before skip=5pt, after skip=5pt,
    boxsep=4pt, top=4pt, bottom=4pt
]
1  

00:00:01,739 $\rightarrow$ 00:00:02,826

那么我是谁?

Then who am I?

\vspace{5pt}

2  

00:00:03,521 $\rightarrow$ 00:00:04,913

你就是那个节目的明星 

You're the star.

\vspace{5pt}

...

\vspace{5pt}
5  

00:00:04,540 $\rightarrow$ 00:00:05,826  

所以才有那么多人看你

That's what made you so good to watch.
\end{tcolorbox}
\end{CJK*}
\caption{An example of a bilingual SRT subtitle file.}
\label{fig:srt}
\end{figure}

It is worth mentioning that the aforementioned post-processing is not necessary for EVE. Due to the fact that taking videos with original frame rate as input will greatly decrease the video duration that EVE can handle, we extract frames of videos in 2-FPS as the input, which requires post-processing to output accurate timestamps. If we only input a two-second video at full frame rate (\textit{e.g.}, 120 frames in total), the model can generate accurate SRT subtitles without the need for post-processing. Therefore, it can be said that EVE is an end-to-end method for video subtitle extraction.

\begin{table}[t]
    \centering
    \caption{The number of samples in different languages. Both short videos and movie videos are mainly in Chinese.}
    \renewcommand{\arraystretch}{1.0}
    \scalebox{1}{
    \begin{tabular}{c|cccc}
    \toprule
    ~ & \textbf{ Chinese} & \textbf{ English }& \textbf{ Bilingual} & \textbf{ None-Text} \\
    \midrule
    
    Short Video & 93,871 & 1,196 & 464 & 9,693 \\
    Movie & 3,141 & 2,600 & 173 & 5,258 \\
    
    \bottomrule
    \end{tabular}}

    \label{tab:num-of-sample}
\end{table}

\section{Dataset}
\label{sec:Dataset}

Although video subtitle extraction has been studied for years, existing datasets for this task remain extremely limited. To this end, we present the ViSa dataset, the first large-scale dataset for video subtitle extraction, providing training material and a evaluation benchmark dataset for this task. In accordance with the previously mentioned training objectives, we divide ViSa into three components: the pre-training dataset, the supervised fine-tuning dataset, and the test dataset.

\subsection{Comparison With Existing Dataset}
Elshahaby \textit{et al.} \cite{elshahaby2022end} introduced the FiViD dataset for video subtitle extraction task, while Yan \textit{et al.} \cite{yan2020end} trained and evaluated their model on publicly available image-based OCR datasets, including MJ Synth \cite{jaderberg2014synthetic}, ICDAR03 \cite{1227749}, ICDAR13 \cite{6628859}, and SVT \cite{Wang20111457}.
In contrast, Mathur \textit{et al.} \cite{mathur2015generating} relied on the audio-only VoxForge \cite{voxforge} corpus to explore speech-to-text subtitle generation, a direction that falls outside the scope of the visual subtitle extraction task studied in this work.
As shown in Table ~\ref{tab:dataset-compare}, the aforementioned datasets are typically limited in scale and lack timestamp annotations. Moreover, most of them composed of images rather than video.
Consequently, these datasets are insufficient to support modern LVLM-based models that require large-scale, high-quality supervision, especially for fine-grained temporal localization.

Meanwhile, recent large video datasets such as CNVid~\cite{gan2023cnvid} and MovieChat-1K~\cite{song2024moviechat} provide massive video–text pairs, yet they are not designed for subtitle extraction.  
They contain video captions or movie subtitles that (1) do not align with frame-level timestamps, (2) are not guaranteed to reflect on-screen subtitle content, and (3) lack start–end boundaries required by this task.  
Thus, they cannot be directly used to train or evaluate a subtitle extraction model.

To fill this gap, we construct  ViSa, the first large-scale dataset specifically designed for timestamped video subtitle extraction.  
ViSa contains bilingual (Chinese and English) subtitle texts paired with both coarse-grained and fine-grained timestamps, enabling scalable pre-training, supervised fine-tuning, and rigorous evaluation of video subtitle extraction systems. Table~\ref{tab:dataset-compare} summarizes the differences between ViSa and prior subtitle extraction datasets.
\begin{table}[t]
\centering
\caption{Comparison of ViSa with datasets used in previous works.}
\renewcommand{\arraystretch}{1.15}
\setlength{\tabcolsep}{4 pt}

\begin{tabular}{c c c >{\centering\arraybackslash}p{1.4cm} c c}
\toprule
\makecell[c]{Dataset} & 
\makecell[c]{Data\\Type} &
\makecell[c]{Scale} &
\makecell[c]{On-screen\\Text Annot.} &
\makecell[c]{Time-\\stamp}  \\
\midrule

FiViD~\cite{elshahaby2022end}
& Image 
& $\approx$74k 
& Yes
& No 
 \\

MJ Synth \cite{jaderberg2014synthetic}
& Synthetic 
& 9M 
& Yes
& No 
\\

ICDAR03  \cite{1227749}
& Image 
& 860
& Yes
& No 
 \\

ICDAR13  \cite{6628859}
& Image 
& 857
& Yes
& No  \\

SVT \cite{Wang20111457}
& Image 
& 647
& Yes
& No 
 \\

CNVid~\cite{gan2023cnvid}
& Video 
& 3.5M 
& No
& No 
 \\

MovieChat-1K~\cite{song2024moviechat}
& Video 
& 1k 
& No
& No 
 \\

\midrule

\textbf{ViSa (ours)} 
& \textbf{Video} 
& \textbf{2.5M}
& \textbf{Yes}
& \textbf{Yes}
 \\

\bottomrule
\end{tabular}
\label{tab:dataset-compare}
\end{table}

\subsection{Data Construction}

The raw videos of ViSa are collected from two datasets: CNVid~\cite{gan2023cnvid} and MovieChat-1K~\cite{song2024moviechat}. CNVid is the largest public video-text dataset in Chinese, which contains 3.5M short videos collect from Douyin~\footnote{https://www.douyin.com/}. MovieChat-1K contains 1K long movie videos with bilingual subtitles.

Based on these two datasets, we construct ViSa for video subtitle extraction. For CNVid, we perform the following three steps to reserve the ideal videos: (1) First, we filter out videos with a duration shorter than 10 seconds or longer than 120 seconds. The reason for this operation is that short videos (duration$<$10s) contain a lot of noise while the number of frames in long videos (duration$>$120s) exceeds the model's input limits. (2) Then, we utilize a in-house video OCR tools to predict the text tracklets in the reserved videos. (3) Finally, videos with less than five tracklets will be filtered out, as these videos are unlikely to contain subtitles. Due to the fact that the movie videos in MovieChat-1K are mostly longer than 30 minutes, we clip each movie video into multiple short videos with a random duration (from 15s to 60s). By processing the above two datasets, we can obtain 2.5M short videos to build the video subtitle extraction dataset ViSa.

\begin{figure*}[t]
  \centering

   \includegraphics[width=0.96\linewidth]{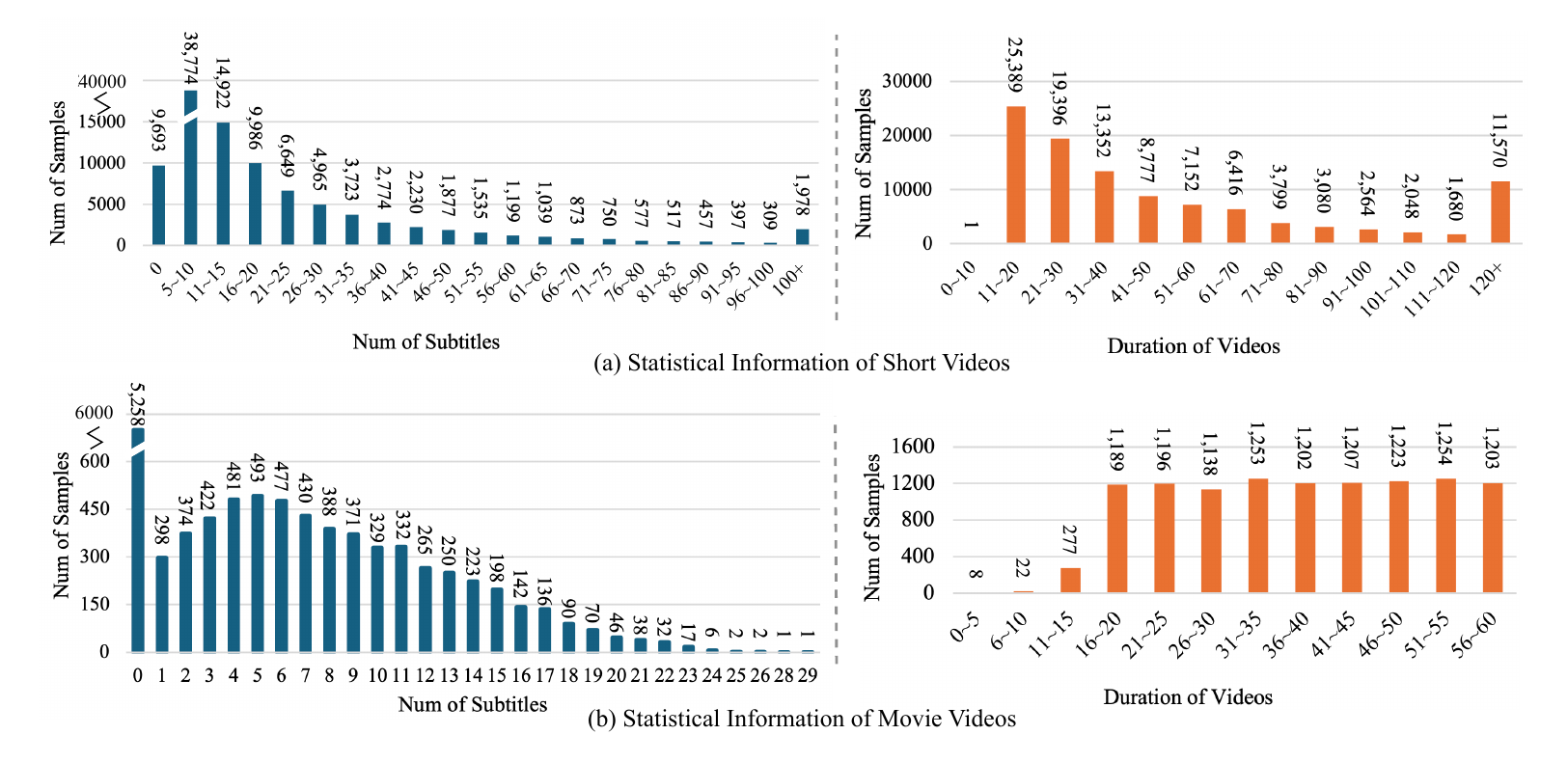}

   \caption{The statistics information of ViSa. The proportion of videos without subtitles in movie videos is higher that in short videos. Since we clip movie videos based on a uniformly sampled duration in the range of 15$-$60 seconds, the duration distribution of movie videos is more balanced compared to short videos.}
   \label{fig:visa}
\end{figure*}

Specifically, ViSa divides the collected videos into three parts: pre-training, supervised fine-tuning, and test datasets. Some statistics of them are described below.

\subsection{Pre-training dataset.} The pre-training dataset contains 2.35M videos, all of which are collected from CNVid. The videos in the pre-training dataset are annotated with text tracking and text recognition since the goal of pre-training is to equip the model with the text perception capability. 
Therefore, this annotation is coarse-grained and low-cost, considering the large scale of this portion of the data.
The majority of pre-training videos are typically shorter than 60 seconds in length. The number of text tracklets in videos mostly ranges between 12 and 30. In a total of 2.35M pre-training videos, 1.8M videos include text tracklets while 0.7M videos contain no texts.

\subsection{Supervised Fine-Tuning dataset.} The videos for Supervised Fine-Tuning  (SFT) originate from both CNVid and MovieChat-1K. Concretely, the SFT dataset includes 210K Chinese videos from CNVid and 20K bilingual videos from MovieChat-1K. Each video in the SFT dataset is annotated with subtitle text tracking and corresponding results. 

Specifically, we use the annotation tool arctime \footnote{https://arctime.org} to perform automated annotation for each video, followed by manual review and correction to ensure the subtitles are accurate and the timestamp error is less than 0.5 seconds.
As a data augmentation strategy to expand the SFT dataset, we annotate each video with both 1-FPS and 2-FPS sampling rate. The statistics of SFT dataset are shown in Figure~\ref{fig:visa}. As shown in Table~\ref{tab:num-of-sample}, among 230K SFT videos, 30K videos does not contain text.

\subsection{Test dataset.} To evaluate the performance of EVE on both short videos and movies, we collect 857 videos from CNVid and 862 videos from MovieChat-1K to build the test dataset. The annotation for this portion of the data is fine-grained and manually generated to enable objective model evaluation. Similar to the SFT dataset, both positive samples (\textit{i.e.}, videos containing texts) and negative samples (\textit{i.e.}, videos containing no texts) are also included in the test dataset.

\begin{table*}[t]
    \centering
        \caption{Performance comparison with existing three types of methods.}
    \renewcommand{\arraystretch}{1.0}
    \scalebox{1.0}{
    \begin{tabular}{cl|cc|c|cc|c}
    \toprule
    \multicolumn{2}{c|}{\multirow{3}*{Method}} & \multicolumn{3}{c|}{\textbf{Short Video}} & \multicolumn{3}{c}{\textbf{Movie}} \\
    ~ & ~ & \multicolumn{2}{c|}{Positive}  & Negative & \multicolumn{2}{c|}{Positive} & Negative \\
    ~ & ~ & \textit{NED} & \textit{SubER} & \textit{SubER} & \textit{NED} & \textit{SubER} & \textit{SubER} \\
    \midrule
    \multirow{3}*{\textit{Open-sourced Tools}} & VideOCR~\cite{videocr} & 0.348 & 72.72 & 52.63 & 0.686 & 6.28 &  57.12 \\
    ~ & VideOCR~\cite{videocr} + Subregion & 0.659 & 52.92 & 43.81 & 0.702 & 5.24 & 66.35 \\
    ~ & PyVideoTrans~\cite{pyvideotrans} & 0.477 & 46.03 & 92.34 & 0.639 & 88.35 & 52.51 \\
    
    \midrule

    \multirow{2}*{\textit{Cloud Service}} & OpenVision~\cite{openvision} & 0.771 & 21.85 & 42.85 & 0.718 & 6.57 & 18.61 \\
    ~ & GhostCut~\cite{ghost} & 0.655 & 54.78 & 46.00 & 0.659 & 7.26 & 24.43\\
    \midrule

    \multirow{3}*{\textit{Open-sourced LVLMs}} & Qwen2-VL-2B~\cite{wang2024qwen2} & 0.068 & - & - & 0.070 & - & - \\
    ~ & Qwen2-VL-7B~\cite{wang2024qwen2} & 0.123 & - & - & 0.227 & - & - \\
    ~ & MiniCPM-V-2.6~\cite{yao2024minicpm} & 0.043 & - & - & 0.047 & - & - \\

    \midrule
    \multirow{2}*{\textit{Ours}} & EVE-0.5B & 0.759 & 22.82 & 18.36 & 0.685  & 5.10 & 17.06   \\
    ~ & EVE-7B & \textbf{0.882}  & \textbf{15.16} & \textbf{13.92}   & \textbf{0.808}  & \textbf{1.64} & \textbf{10.15} \\
    
    \bottomrule
    \end{tabular}}

    \label{cmp-table}
\end{table*}

\section{Experiments}
\label{exp}

\subsection{Implementation Details}
The proposed method is implemented with PyTorch. All experiments are conducted on $24$ A100 GPUs. The proposed EVE is trained using the AdamW optimizer~\cite{loshchilov2017decoupled} where the learning rate is set to $5\times 10^{-5}$. The input frames are resized into $640\times640$. The batch size is set to 48 for pre-training and 16 for supervised fine-tuning. The shape of $\mathbf{F}^v_i$ is set to $\mathbb{R}^{4\times 4\times C}$. The number of queries $Q$ in the TSTQ is set to 10, and the number of visual tokens after SSCA is set to 16.  The vision encoder is initialized with the pre-trained parameters of Donut~\cite{kim2021donut}, and the large language model loads the pre-trained models of Qwen2-7B or Qwen2-0.5B.

\subsection{Metrics}
In this paper, we adopt two types of metrics to evaluate the performance of models: Normalized Edit Distance (NED)~\cite{zhang2019icdar} and SubER~\cite{wilken2022suber}. NED is defined as:
\begin{equation}
    \texttt{NED}(\hat{\mathbf{y}}_i, \mathbf{y}_i) = 1 - \frac{1}{S} \sum^S_{i=1} \texttt{ED}(\hat{\mathbf{y}}_i, \mathbf{y}_i)/\texttt{MaxL}(\hat{\mathbf{y}}_i, \mathbf{y}_i)
\end{equation}
where \texttt{ED}($\cdot$) denotes the edit distance function. The function \texttt{MaxL}(\texttt{A}, \texttt{B}) returns the maximum length between strings \texttt{A} and \texttt{B}. $S$ represents the number of samples. The NED metric is utilized to calculate the accuracy of predicting subtitle texts. However, it may hardly evaluate the accuracy of timestamp predictions. Therefore, we additionally adopt SubER~\cite{wilken2022suber} to evaluate the performance of models. SubER is defined as:
\begin{equation}
    \texttt{SUBER} = \frac{\# \text{word}\ \ \text{edits} + \# \text{break} \ \ \text{edits} + \# \text{shifts}}{\# \text{reference} \ \  \text{words} + \# \text{reference} \ \ \text{breaks}}
\end{equation}
where ``\#'' means ``number of''. \textbf{word edits} are insertions, deletions and substitutions of words; \textbf{break edits} are insertions, deletions and substitutions of line breaks; \textbf{shifts} are movements of one or more adjacent hypothesis tokens to a position of a matching phrase in the reference; \textbf{reference words}/\textbf{reference breaks} is the ground-truth words/breaks. More details about SubER can refer to~\cite{wilken2022suber}.

\begin{table}[t]
    \centering
    \caption{Performance comparison with two typical token compression methods.}
    \renewcommand{\arraystretch}{1.05}
    \setlength{\tabcolsep}{4.3pt}
    \begin{tabular}{l|cc|c|cc|c}
    \toprule
    \multicolumn{1}{c|}{\multirow{3}{*}{\textbf{Method}}} & 
    \multicolumn{3}{c|}{\textbf{Short Video}} & 
    \multicolumn{3}{c}{\textbf{Movie}} \\
     ~ & \multicolumn{2}{c|}{Positive}  & Negative & 
     \multicolumn{2}{c|}{Positive} & Negative \\
     ~ & \textit{NED} & \textit{SubER} & \textit{SubER} & 
     \textit{NED} & \textit{SubER} & \textit{SubER} \\
    \midrule
    T-Selector~\cite{wang2025elysium} & 0.801 & 20.32 & 23.84 & 0.731 & 8.72 & 19.30 \\
    ToMe~\cite{bolyatoken} & 0.714 & 28.35 & 25.62 & 0.705 & 10.79 & 15.59 \\
    \midrule
    EVE-0.5B & 0.759 & 22.82 & 18.36 & 0.685 & 5.10 & 17.06 \\
    EVE-7B & \textbf{0.882} & \textbf{15.16} & \textbf{13.92} & 
              \textbf{0.808} & \textbf{1.64} & \textbf{10.15} \\
    \bottomrule
    \end{tabular}
    \label{cmp-supp}
\end{table}

\subsection{Experimental Results}

\subsubsection{Evaluating Existing Methods} We evaluated three categories of video subtitle extraction methods on the proposed test dataset as shown in Table~ \ref{cmp-table}. Among the open-sourced tools, PyVideoTrans based on Automatic Speech Recognition (ASR) achieves relatively inferior performance on negative samples compared with VideOCR. The reason for these results is that ASR-based methods inherently treat spoken audio as potential subtitles,
causing false positives in negative samples where no on-screen text exists. We also evaluated two cloud services that can perform video subtitle extraction. GhostCut cannot perform well on short videos containing a lot of background texts since it tends to mistake background texts as subtitles. In contrast, OpenVision can achieve relatively ideal performance on both short videos and movie videos. Two open-source multi-modal large models (Qwen2-VL~\cite{wang2024qwen2} and MiniCPM-V-2.6~\cite{yao2024minicpm}) perform poorly on the video subtitle extraction task because they are not trained on corresponding datasets. They prefer to performing video captioning rather than extracting subtitle texts, thus resulting in subpar performance. In addition, the experimental results demonstrate that Qwen2-VL's video OCR capability is better than MiniCPM-V-2.6.

\begin{table}[t]
    \centering
    \caption{Ablation studies on the proposed EVE (\textit{metric: SubER}). }
    \renewcommand{\arraystretch}{1.05} 
    \setlength{\tabcolsep}{8.5pt} 
    \begin{tabular}{ccc|cc|cc}
    \toprule
    \multirow{2}{*}{\textbf{SSCA}} & 
    \multirow{2}{*}{\textbf{TSTQ}} & 
    \multirow{2}{*}{\textbf{SP}} & 
    \multicolumn{2}{c|}{\textbf{Short Video}} & 
    \multicolumn{2}{c}{\textbf{Movie}} \\
    ~ & ~ & ~ & \textbf{Pos} & \textbf{Neg} & \textbf{Pos} & \textbf{Neg} \\
    \midrule
    \ding{51} & ~ & ~ & 18.15 & 19.34 & 7.25 & 14.26 \\
    ~ & \ding{51} & ~ & 22.35 & 20.19 & 9.41 & 17.80 \\
    \ding{51} & \ding{51} & ~ & 17.32 & 16.28 & 4.17 & 15.82 \\
    \ding{51} & \ding{51} & \ding{51} & 
    \textbf{15.16} & \textbf{13.92} & 
    \textbf{1.64} & \textbf{10.15} \\
    \bottomrule
    \end{tabular}
    \label{ablation}
\end{table}

\subsubsection{Comparison with Existing Methods} In this paper, we establish two models with different LLMs. To achieve better performance in video subtitle extraction, we utilize Qwen2-7B to construct EVE-7B. Through the experimental results in Table~\ref{cmp-table}, EVE-7B achieves the best performance compared with three types of video subtitle extraction methods. For example, on the short video dataset, EVE-7B outperforms OpenVision by 11.1\% in the metric of NED. Considering that the EVE-7B will cost more computing resources in practical applications, we establish a light-weighted model EVE-0.5B with Qwen2-0.5B. The experimental results in Table~\ref{cmp-table} demonstrate that despite using the same dataset for training, EVE-0.5B still cannot achieve the same performance as EVE-7B. However, benefiting from the proposed framework and training strategy, it can still surpass most of existing methods.
\begin{figure*}[htbp]
  \centering

   \includegraphics[width=0.94\linewidth]{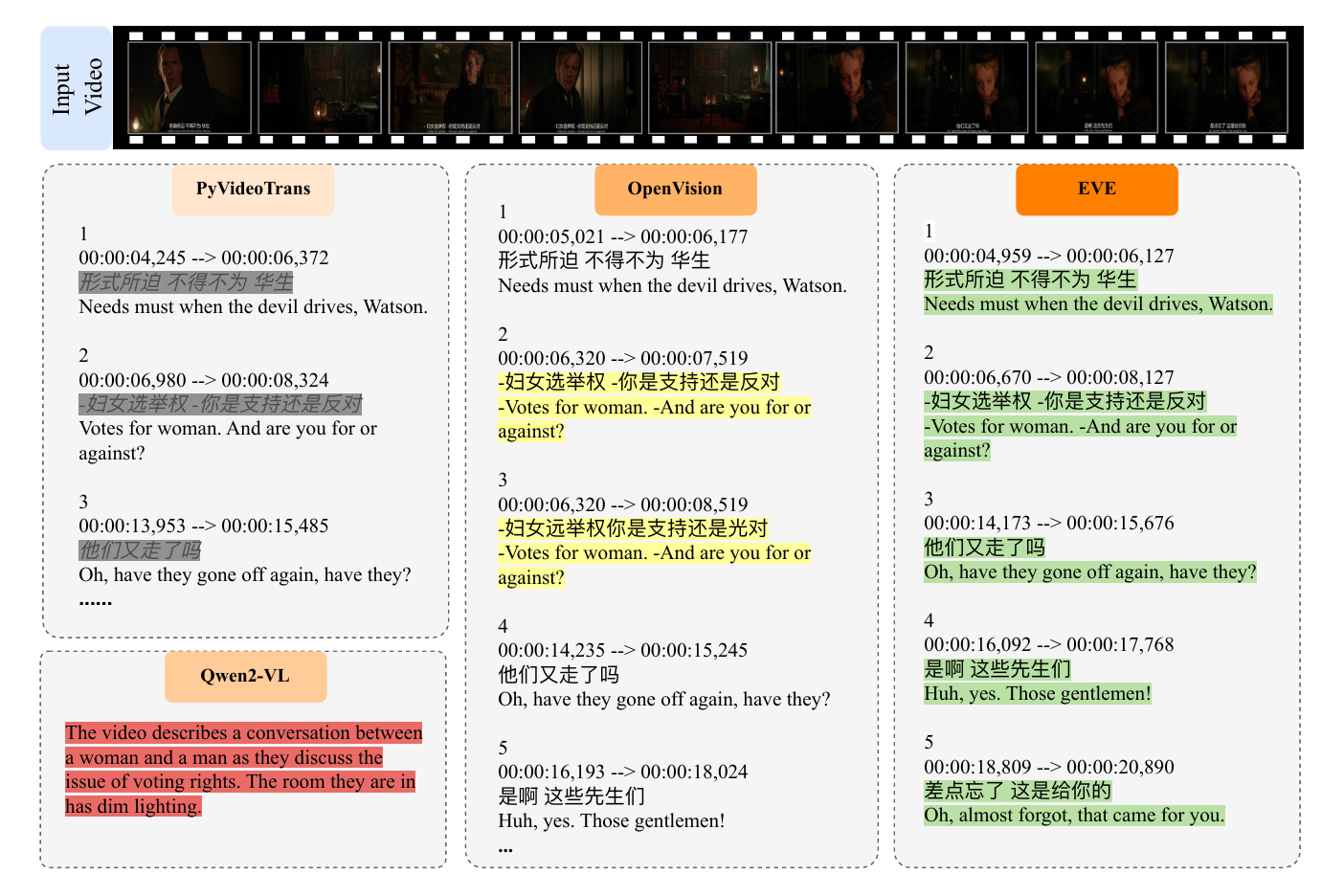}

   \caption{An example of subtitle extraction for different methods. \colorbox{gray}{Gray} and \colorbox{mycolor2}{Yellow} indicate the missing and repetition prediction problem, respectively. \colorbox{mycolor}{Red} represents the format of the answer is incorrect. \colorbox{mycolor1}{Green} represent the correct prediction content.}
   \label{fig:res}
\end{figure*}

\begin{figure}[ht]
   \begin{center}
   \includegraphics[width=0.42\textwidth]{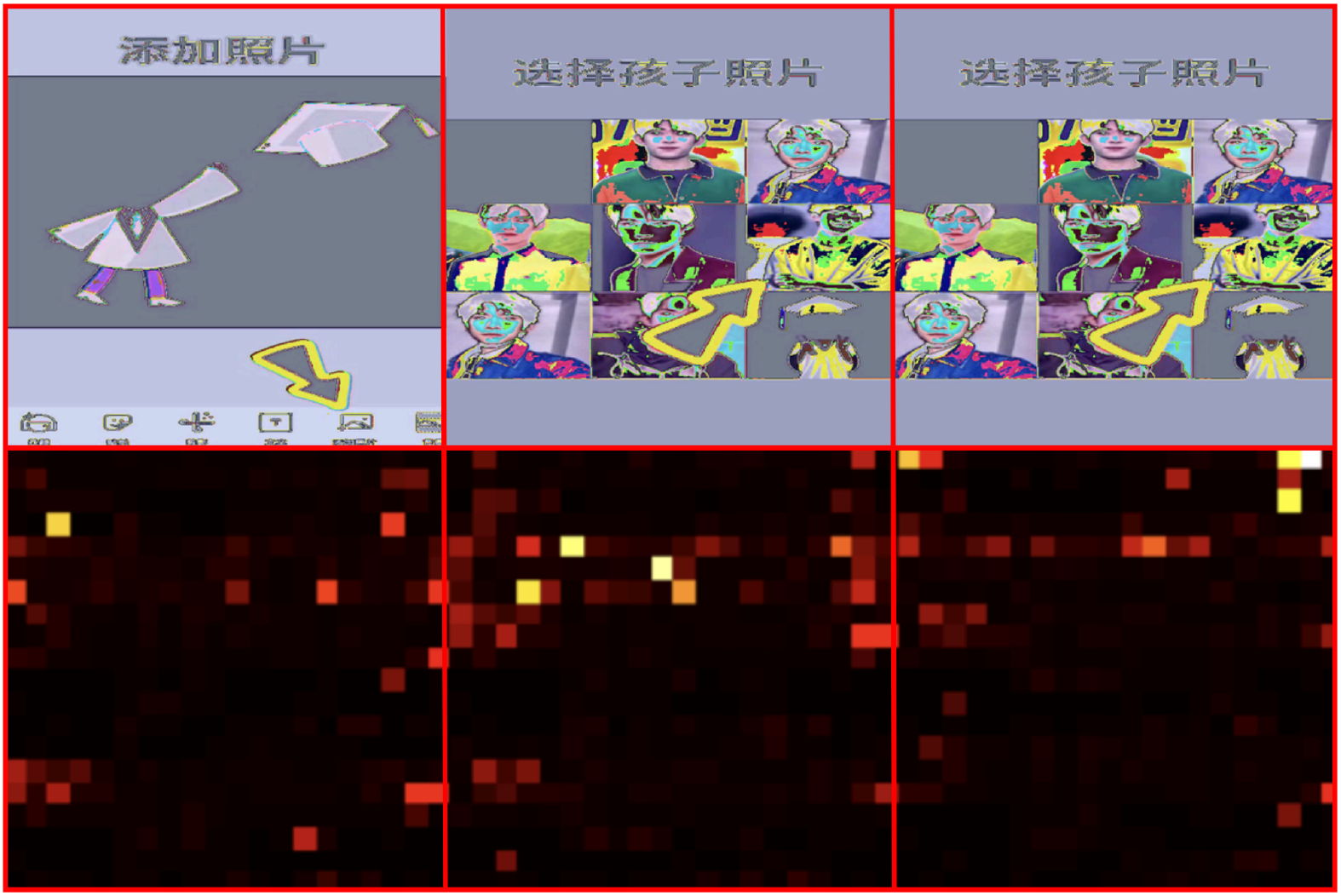}
   \end{center}
    \centering
   \caption{The visualization for the temporal subtitle token querying using heatmap.}
   \label{fig:vis}
\end{figure}

\begin{figure*}[htbp]
   \begin{center}
   \includegraphics[width=0.96\textwidth]{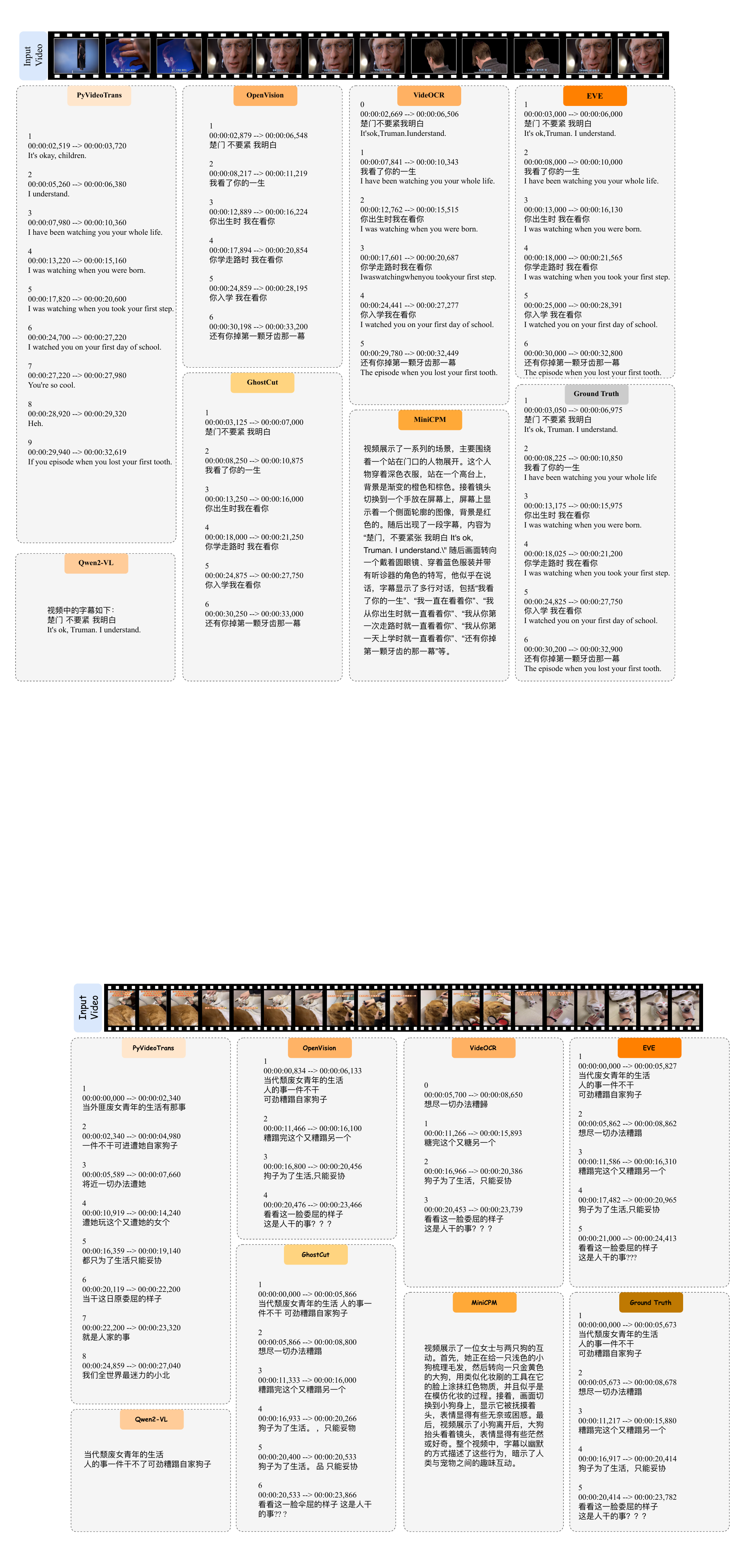}
   \end{center}
   \caption{Visualization of subtitle extraction results with timestamps from a movie clip using MiniCPM, GhostCut, VideOCR, Qwen2-VL, and the proposed EVE. }
   \label{fig:vis_movie}
\end{figure*}

\subsection{Comparison with Token Compression Methods}
Beyond comparisons with existing subtitle extraction tools and LVLM-based baselines, we further evaluate the proposed S\textsuperscript{3} module against two representative token compression approaches, T-Selector~\cite{wang2025elysium} and ToMe~\cite{bolyatoken}, as summarized in Table~\ref{cmp-supp}.
As mentioned in the Related Work section, these methods are functionally similar to the proposed S\textsuperscript{3} module, as they all aim to extract key information to feed into the LLM. Therefore, we conduct this experiment for comparison.
Both methods are re-implemented using the Qwen2-7B backbone to ensure a fair comparison under identical conditions. The results reveal that despite utilizing a smaller 0.5B language model, EVE-0.5B achieves performance on par with or even surpasses these token compression baselines, validating the efficiency of our design. When scaled to EVE-7B, the performance improves substantially across all benchmarks. For instance, compared to ToMe, EVE achieves a 16.8\% improvement in NED on the Short Videos test set, demonstrating stronger temporal alignment and finer-grained token utilization. This consistent advantage highlights the effectiveness of our proposed S\textsuperscript{3} module, which interleaves vision and text representations to selectively preserve semantically relevant tokens while filtering out redundant ones. Unlike conventional compression schemes that rely purely on token similarity, S\textsuperscript{3} module adaptively integrates multimodal cues, enabling more precise identification of subtitle-relevant regions and enhancing overall extraction robustness across both short and long-form videos.

\subsection{Ablation Studies}

In this paper, we propose an innovative adapter called S\textsuperscript{3} module, which includes a Spatial Semantic Context Aggregate (SSCA) and a Temporal Subtitle Token Query (TSTQ). Additionally, we introduce the separate projector (SP) to project the  global  visual tokens and text-related tokens into the input space of large language models. In this section, we conduct ablation studies on these three modules. The experimental results, as shown in Table~\ref{ablation}, indicate that introducing either the SSCA or the TSTQ alone does not outperform previous methods. However, when both of them are integrated into S\textsuperscript{3} module, there is a significant improvement in model performance. Compared to only using the TSTQ, our method achieves a 5.03\% and 5.28\% improvement in SubER metrics on the positive samples of short video and movie datasets, respectively. Furthermore, introducing separate projectors can further improve the performance. A plausible explanation is that the two types of information focus on different aspects of visual effects, and thus, separate projectors facilitate better alignment of features with LLMs.

\subsection{Hyperparameter Selection}

There are two hyperparameters in EVE: the number of queries in the TSTQ and the number of visual tokens after SSCA. We have conducted corresponding experiments with different numbers of tokens. When the number of visual tokens (denoted as $N$) and the number of queries in the TSTQ (denoted as $K$) are set to larger values of ($N$=36 and $K$=20), the performance of EVE is only improved by 0.3\% in the NED metric, but the computational cost will increase significantly, due to more tokens being reserved for the LLM. When $N$ and $K$ are set to smaller values ($N$=4 and $K$=1), there is a significant decrease in the performance of EVE, with the NED metric decreasing by nearly 28.6\%. In order to balance computational cost and model performance, we ultimately set $N$ and $K$ to 16 and 10, respectively.

\subsection{Visualization for Temporal Subtitle Token Query}

To better understand the role of the Temporal Subtitle Token Query in EVE, we visualize its attention responses across representative frames, as shown in Figure~\ref{fig:vis}. The upper row depicts the original video frames, while the lower row illustrates the corresponding attention heatmaps. We observe that the TSTQ consistently activates in regions containing textual elements, such as interface buttons, titles, and overlaid text cues. This selective activation indicates that the module effectively distinguishes text-related areas from irrelevant visual content, enabling more efficient information compression and representation. Moreover, as the video progresses, the attention maps dynamically adjust to newly appearing text, demonstrating strong adaptability to temporal variations. These visualizations validate that the TSTQ successfully enhances the model’s perception of text-bearing regions and contributes to more precise multimodal understanding within EVE.

\subsection{Visualization for Video Subtitle Extraction}
In Figure~\ref{fig:res}, we compare PyVideoTrans \cite{pyvideotrans}, OpenVision \cite{openvision}, Qwen2-VL \cite{wang2024qwen2}, and the proposed EVE for subtitle extraction. PyVideoTrans omits critical dialogue segments, while OpenVision introduces redundancy through repeated outputs. Qwen2-VL diverges from the task objective, producing scene-level descriptions rather than precise transcripts. By contrast, EVE generates temporally aligned, semantically faithful, and duplication-free subtitles. For example, the utterance “Oh, have they gone off again, have they?” is recovered accurately and uniquely. These results confirm EVE’s robustness and highlight its effectiveness as a reliable framework for multimodal subtitle generation.

In Figure~\ref{fig:vis_movie}, we further compare MiniCPM ~\cite{yao2024minicpm}, GhostCut \cite{ghost}, VideOCR \cite{videocr}, Qwen2-VL \cite{wang2024qwen2}, and EVE for video subtitle extraction. MiniCPM fails to perform transcription and instead outputs verbose scene-level descriptions. GhostCut captures most subtitle lines but exhibits minor temporal misalignment and redundant repetitions. VideOCR shows high recognition quality but suffers from token concatenation issues such as “It’s ok, Truman. I understand.” caused by inaccurate text segmentation. Qwen2-VL generates only a single incomplete line, failing to cover the entire dialogue sequence. In contrast, EVE delivers temporally consistent, structurally complete, and duplication-free subtitles. For instance, the utterance “It’s ok, Truman. I understand.” is precisely aligned and clearly formatted, demonstrating EVE’s strong capability in maintaining accurate temporal correspondence and structural completeness.  Moreover, the visualization results indicate that the timestamps predicted by EVE are closer to the ground truth compared to other methods.

\section{Conclusion}
\label{Conclusion}
In this paper, we propose an End-to-end Video Subtitle Extraction model, called EVE, which consists of a vision encoder, an adapter module and a large language model. In EVE, a novel adapter called the S\textsuperscript{3} module has been introduced, which is able to model salient subtitle representations while retaining only a minimal number of tokens. To benchmark the video subtitle extraction task, we propose a large bilingual (\textit{i.e.}, Chinese and English) dataset ViSa, including 2.35M videos for pre-training, 230K videos for fine-tuning and 1.6K videos for test. Through experimental results, we observe that the proposed EVE achieves the best performance in both NED and SubER compared with previous methods for video subtitle extraction. Despite its strong performance, EVE still faces two limitations. 
First, although each frame is compressed into only 26 tokens, the maximum input length remains insufficient for very long videos, such as those exceeding one hour. 
Second, EVE predicts only subtitle texts and timestamps, without providing spatial coordinates of subtitle regions. 
Future work will explore more efficient token compression strategies to support longer input sequences and extend EVE to localize subtitle positions for broader downstream applications.

\bibliographystyle{IEEEtran}
\bibliography{main}

@ARTICLE{tip-video-sum1,
  author={Li, Wenrui and Han, Wei and Deng, Liang-Jian and Xiong, Ruiqin and Fan, Xiaopeng},
  journal={IEEE Transactions on Image Processing}, 
  title={Spiking Variational Graph Representation Inference for Video Summarization}, 
  year={2025},
  volume={34},
  number={},
  pages={5697-5709},
  doi={10.1109/TIP.2025.3602649},
  ISSN={1941-0042},
  month={},}

@ARTICLE{tip-video-sum2,
  author={Ye, Cheng and Chen, Weidong and Hu, Bo and Zhang, Lei and Zhang, Yongdong and Mao, Zhendong},
  journal={IEEE Transactions on Image Processing}, 
  title={Improving Video Summarization by Exploring the Coherence Between Corresponding Captions}, 
  year={2025},
  volume={34},
  number={},
  pages={5369-5384},
  doi={10.1109/TIP.2025.3598709}}

@ARTICLE{tip-video-ret,
  author={Wen, Jun and Chen, Yufeng and Shi, Ruiqi and Ji, Wei and Yang, Menglin and Gao, Difei and Yuan, Junsong and Zimmermann, Roger},
  journal={IEEE Transactions on Image Processing}, 
  title={HOVER: Hyperbolic Video-Text Retrieval}, 
  year={2025},
  volume={34},
  number={},
  pages={6192-6203},
  doi={10.1109/TIP.2025.3611174}}

@String(VR   = {Vis. Res.})

@inproceedings{wu2021towards,
  title={Towards long-form video understanding},
  author={Wu, Chao-Yuan and Krahenbuhl, Philipp},
  booktitle={Proceedings of the IEEE/CVF Conference on Computer Vision and Pattern Recognition},
  pages={1884--1894},
  year={2021}
}

@incollection{wu2024deep,
  title={Deep Learning Basics for Video Understanding},
  author={Wu, Zuxuan and Jiang, Yu-Gang},
  booktitle={Deep Learning for Video Understanding},
  pages={7--20},
  year={2024},
  publisher={Springer}
}

@article{elshahaby2022end,
  title={An end to end system for subtitle text extraction from movie videos},
  author={Elshahaby, Hossam and Rashwan, Mohsen},
  journal={Journal of Ambient Intelligence and Humanized Computing},
  volume={13},
  number={4},
  pages={1853--1865},
  year={2022},
  publisher={Springer}
}

@article{yao2024deco,
  title={DeCo: Decoupling Token Compression from Semantic Abstraction in Multimodal Large Language Models},
  author={Yao, Linli and Li, Lei and Ren, Shuhuai and Wang, Lean and Liu, Yuanxin and Sun, Xu and Hou, Lu},
  journal={arXiv preprint arXiv:2405.20985},
  year={2024}
}

@article{wilken2022suber,
  title={SubER: A Metric for Automatic Evaluation of Subtitle Quality},
  author={Wilken, Patrick and Georgakopoulou, Panayota and Matusov, Evgeny},
  journal={arXiv preprint arXiv:2205.05805},
  year={2022}
}

@inproceedings{gan2023cnvid,
  title={Cnvid-3.5 m: Build, filter, and pre-train the large-scale public chinese video-text dataset},
  author={Gan, Tian and Wang, Qing and Dong, Xingning and Ren, Xiangyuan and Nie, Liqiang and Guo, Qingpei},
  booktitle={Proceedings of the IEEE/CVF Conference on Computer Vision and Pattern Recognition},
  pages={14815--14824},
  year={2023}
}

@inproceedings{song2024moviechat,
  title={Moviechat: From dense token to sparse memory for long video understanding},
  author={Song, Enxin and Chai, Wenhao and Wang, Guanhong and Zhang, Yucheng and Zhou, Haoyang and Wu, Feiyang and Chi, Haozhe and Guo, Xun and Ye, Tian and Zhang, Yanting and others},
  booktitle={Proceedings of the IEEE/CVF Conference on Computer Vision and Pattern Recognition},
  pages={18221--18232},
  year={2024}
}

@article{kim2024image,
  title={An image grid can be worth a video: Zero-shot video question answering using a vlm},
  author={Kim, Wonkyun and Choi, Changin and Lee, Wonseok and Rhee, Wonjong},
  journal={arXiv preprint arXiv:2403.18406},
  year={2024}
}

@article{wang2024videotree,
  title={VideoTree: Adaptive Tree-based Video Representation for LLM Reasoning on Long Videos},
  author={Wang, Ziyang and Yu, Shoubin and Stengel-Eskin, Elias and Yoon, Jaehong and Cheng, Feng and Bertasius, Gedas and Bansal, Mohit},
  journal={arXiv preprint arXiv:2405.19209},
  year={2024}
}

@article{xu2024pllava,
  title={Pllava: Parameter-free llava extension from images to videos for video dense captioning},
  author={Xu, Lin and Zhao, Yilin and Zhou, Daquan and Lin, Zhijie and Ng, See Kiong and Feng, Jiashi},
  journal={arXiv preprint arXiv:2404.16994},
  year={2024}
}

@inproceedings{wu2023bidirectional,
  title={Bidirectional cross-modal knowledge exploration for video recognition with pre-trained vision-language models},
  author={Wu, Wenhao and Wang, Xiaohan and Luo, Haipeng and Wang, Jingdong and Yang, Yi and Ouyang, Wanli},
  booktitle={Proceedings of the IEEE/CVF conference on computer vision and pattern recognition},
  pages={6620--6630},
  year={2023}
}

@inproceedings{liu2021swin,
  title={Swin transformer: Hierarchical vision transformer using shifted windows},
  author={Liu, Ze and Lin, Yutong and Cao, Yue and Hu, Han and Wei, Yixuan and Zhang, Zheng and Lin, Stephen and Guo, Baining},
  booktitle={Proceedings of the IEEE/CVF international conference on computer vision},
  pages={10012--10022},
  year={2021}
}

@article{yang2024qwen2,
  title={Qwen2 technical report},
  author={Yang, An and Yang, Baosong and Hui, Binyuan and Zheng, Bo and Yu, Bowen and Zhou, Chang and Li, Chengpeng and Li, Chengyuan and Liu, Dayiheng and Huang, Fei and others},
  journal={arXiv preprint arXiv:2407.10671},
  year={2024}
}

@article{wang2024qwen2,
  title={Qwen2-vl: Enhancing vision-language model's perception of the world at any resolution},
  author={Wang, Peng and Bai, Shuai and Tan, Sinan and Wang, Shijie and Fan, Zhihao and Bai, Jinze and Chen, Keqin and Liu, Xuejing and Wang, Jialin and Ge, Wenbin and others},
  journal={arXiv preprint arXiv:2409.12191},
  year={2024}
}

@article{yan2020end,
  title={End-to-end video subtitle recognition via a deep residual neural network},
  author={Yan, Hongyu and Xu, Xin},
  journal={Pattern Recognition Letters},
  volume={131},
  pages={368--375},
  year={2020},
  publisher={Elsevier}
}

@inproceedings{tian2016detecting,
  title={Detecting text in natural image with connectionist text proposal network},
  author={Tian, Zhi and Huang, Weilin and He, Tong and He, Pan and Qiao, Yu},
  booktitle={Computer Vision--ECCV 2016: 14th European Conference, Amsterdam, The Netherlands, October 11-14, 2016, Proceedings, Part VIII 14},
  pages={56--72},
  year={2016},
  organization={Springer}
}

@inproceedings{ramani2020automatic,
  title={Automatic subtitle generation for videos},
  author={Ramani, Aditya and Rao, Asmita and Vidya, V and Prasad, VR Badri},
  booktitle={2020 6th International Conference on Advanced Computing and Communication Systems (ICACCS)},
  pages={132--135},
  year={2020},
  organization={IEEE}
}

@inproceedings{bolyatoken,
  title={Token Merging: Your ViT But Faster},
  author={Bolya, Daniel and Fu, Cheng-Yang and Dai, Xiaoliang and Zhang, Peizhao and Feichtenhofer, Christoph and Hoffman, Judy},
  booktitle={The Eleventh International Conference on Learning Representations},
    year={2022}
}

@article{liang2022not,
  title={Not all patches are what you need: Expediting vision transformers via token reorganizations},
  author={Liang, Youwei and Ge, Chongjian and Tong, Zhan and Song, Yibing and Wang, Jue and Xie, Pengtao},
  journal={arXiv preprint arXiv:2202.07800},
  year={2022}
}

@inproceedings{liusimple,
  title={A Simple Romance Between Multi-Exit Vision Transformer and Token Reduction},
  author={Liu, Dongyang and Kan, Meina and Shan, Shiguang and Xilin, CHEN},
  booktitle={The Twelfth International Conference on Learning Representations},
    year={2024}
}

@inproceedings{li2023blip,
  title={Blip-2: Bootstrapping language-image pre-training with frozen image encoders and large language models},
  author={Li, Junnan and Li, Dongxu and Savarese, Silvio and Hoi, Steven},
  booktitle={International conference on machine learning},
  pages={19730--19742},
  year={2023},
  organization={PMLR}
}

@inproceedings{li2025llama,
  title={Llama-vid: An image is worth 2 tokens in large language models},
  author={Li, Yanwei and Wang, Chengyao and Jia, Jiaya},
  booktitle={European Conference on Computer Vision},
  pages={323--340},
  year={2025},
  organization={Springer}
}

@inproceedings{ren2024timechat,
  title={Timechat: A time-sensitive multimodal large language model for long video understanding},
  author={Ren, Shuhuai and Yao, Linli and Li, Shicheng and Sun, Xu and Hou, Lu},
  booktitle={Proceedings of the IEEE/CVF Conference on Computer Vision and Pattern Recognition},
  pages={14313--14323},
  year={2024}
}

@article{zhang2023video,
  title={Video-llama: An instruction-tuned audio-visual language model for video understanding},
  author={Zhang, Hang and Li, Xin and Bing, Lidong},
  journal={arXiv preprint arXiv:2306.02858},
  year={2023}
}

@inproceedings{huang2025lita,
  title={Lita: Language instructed temporal-localization assistant},
  author={Huang, De-An and Liao, Shijia and Radhakrishnan, Subhashree and Yin, Hongxu and Molchanov, Pavlo and Yu, Zhiding and Kautz, Jan},
  booktitle={European Conference on Computer Vision},
  pages={202--218},
  year={2025},
  organization={Springer}
}

@article{maaz2023video,
  title={Video-chatgpt: Towards detailed video understanding via large vision and language models},
  author={Maaz, Muhammad and Rasheed, Hanoona and Khan, Salman and Khan, Fahad Shahbaz},
  journal={arXiv preprint arXiv:2306.05424},
  year={2023}
}

@article{liu2024oryx,
  title={Oryx MLLM: On-Demand Spatial-Temporal Understanding at Arbitrary Resolution},
  author={Liu, Zuyan and Dong, Yuhao and Liu, Ziwei and Hu, Winston and Lu, Jiwen and Rao, Yongming},
  journal={arXiv e-prints},
  pages={arXiv--2409},
  year={2024}
}

@article{dosovitskiy2020image,
  title={An image is worth 16x16 words: Transformers for image recognition at scale},
  author={Dosovitskiy, Alexey},
  journal={arXiv preprint arXiv:2010.11929},
  year={2020}
}

@article{rao2021dynamicvit,
  title={Dynamicvit: Efficient vision transformers with dynamic token sparsification},
  author={Rao, Yongming and Zhao, Wenliang and Liu, Benlin and Lu, Jiwen and Zhou, Jie and Hsieh, Cho-Jui},
  journal={Advances in neural information processing systems},
  volume={34},
  pages={13937--13949},
  year={2021}
}

@article{kim2021donut,
  title={Donut: Document understanding transformer without ocr},
  author={Kim, Geewook and Hong, Teakgyu and Yim, Moonbin and Park, Jinyoung and Yim, Jinyeong and Hwang, Wonseok and Yun, Sangdoo and Han, Dongyoon and Park, Seunghyun},
  journal={arXiv preprint arXiv:2111.15664},
  volume={7},
  number={15},
  pages={2},
  year={2021}
}

@article{touvron2023llama,
  title={Llama 2: Open foundation and fine-tuned chat models},
  author={Touvron, Hugo and Martin, Louis and Stone, Kevin and Albert, Peter and Almahairi, Amjad and Babaei, Yasmine and Bashlykov, Nikolay and Batra, Soumya and Bhargava, Prajjwal and Bhosale, Shruti and others},
  journal={arXiv preprint arXiv:2307.09288},
  year={2023}
}

@article{loshchilov2017decoupled,
  title={Decoupled weight decay regularization},
  author={Loshchilov, I},
  journal={arXiv preprint arXiv:1711.05101},
  year={2017}
}

@inproceedings{zhang2019icdar,
  title={Icdar 2019 robust reading challenge on reading chinese text on signboard},
  author={Zhang, Rui and Zhou, Yongsheng and Jiang, Qianyi and Song, Qi and Li, Nan and Zhou, Kai and Wang, Lei and Wang, Dong and Liao, Minghui and Yang, Mingkun and others},
  booktitle={2019 international conference on document analysis and recognition (ICDAR)},
  pages={1577--1581},
  year={2019},
  organization={IEEE}
}

@article{yao2024minicpm,
  title={Minicpm-v: A gpt-4v level mllm on your phone},
  author={Yao, Yuan and Yu, Tianyu and Zhang, Ao and Wang, Chongyi and Cui, Junbo and Zhu, Hongji and Cai, Tianchi and Li, Haoyu and Zhao, Weilin and He, Zhihui and others},
  journal={arXiv preprint arXiv:2408.01800},
  year={2024}
}

@misc{videocr,
    author = "",
    title = "",
    howpublished = "\url{https://github.com/devmaxxing/videocr-PaddleOCR}",
    note = "",
    year = {}
}

@misc{pyvideotrans,
    author = "",
    title = "",
    howpublished = "\url{https://github.com/jianchang512/pyvideotrans}",
    note = "",
    year = {}
}

@article{jaderberg2014synthetic,
  title={Synthetic data and artificial neural networks for natural scene text recognition},
  author={Jaderberg, Max and Simonyan, Karen and Vedaldi, Andrea and Zisserman, Andrew},
  journal={arXiv preprint arXiv:1406.2227},
  year={2014}
}

@INPROCEEDINGS{1227749,
  author={Lucas, S.M. and Panaretos, A. and Sosa, L. and Tang, A. and Wong, S. and Young, R.},
  booktitle={Seventh International Conference on Document Analysis and Recognition, 2003. Proceedings.}, 
  title={ICDAR 2003 robust reading competitions}, 
  year={2003},
  volume={},
  number={},
  pages={682-687},
  doi={10.1109/ICDAR.2003.1227749}}

@ONLINE{voxforge,
  author = {VoxForge Community},
  title  = {VoxForge -- Free Speech Corpus},
  year   = {2025},
  url    = {http://www.voxforge.org/}
}

@INPROCEEDINGS{6628859,
  author={Karatzas, Dimosthenis and Shafait, Faisal and Uchida, Seiichi and Iwamura, Masakazu and Bigorda, Lluis Gomez i and Mestre, Sergi Robles and Mas, Joan and Mota, David Fernandez and Almazàn, Jon Almazàn and de las Heras, Lluís Pere},
  booktitle={2013 12th International Conference on Document Analysis and Recognition}, 
  title={ICDAR 2013 Robust Reading Competition}, 
  year={2013},
  volume={},
  number={},
  pages={1484-1493},
  doi={10.1109/ICDAR.2013.221}}

@CONFERENCE{Wang20111457,
	author = {Wang, Kai and Babenko, Boris and Belongie, Serge},
	title = {End-to-end scene text recognition},
	year = {2011},
	journal = {Proceedings of the IEEE International Conference on Computer Vision},
	pages = {1457 – 1464},
	doi = {10.1109/ICCV.2011.6126402},
	url = {https://www.scopus.com/inward/record.uri?eid=2-s2.0-84863057818&doi=10.1109%2fICCV.2011.6126402&partnerID=40&md5=126084f315916c6397d9e28b9191e3a2},
	type = {Conference paper},
	publication_stage = {Final},
	source = {Scopus},
	note = {Cited by: 1166}
}

@misc{openvision,
    author = "",
    title = "",
    howpublished = "\url{https://vision.aliyun.com/}",
    note = "",
    year = {}
}

@misc{ghost,
    author = "",
    title = "",
    howpublished = "\url{https://cn.jollytoday.com/home/}",
    note = "",
    year = {}
}

@article{papi2023direct,
  title={Direct speech translation for automatic subtitling},
  author={Papi, Sara and Gaido, Marco and Karakanta, Alina and Cettolo, Mauro and Negri, Matteo and Turchi, Marco},
  journal={Transactions of the Association for Computational Linguistics},
  volume={11},
  pages={1355--1376},
  year={2023},
  publisher={MIT Press One Broadway, 12th Floor, Cambridge, Massachusetts 02142, USA~…}
}

@inproceedings{mathur2015generating,
  title={Generating subtitles automatically using audio extraction and speech recognition},
  author={Mathur, Abhinav and Saxena, Tanya and Krishnamurthi, Rajalakshmi},
  booktitle={2015 IEEE International Conference on Computational Intelligence \& Communication Technology},
  pages={621--626},
  year={2015},
  organization={IEEE}
}

@inproceedings{wang2025elysium,
  title={Elysium: Exploring object-level perception in videos via mllm},
  author={Wang, Han and Ye, Yongjie and Wang, Yanjie and Nie, Yuxiang and Huang, Can},
  booktitle={European Conference on Computer Vision},
  pages={166--185},
  year={2025},
  organization={Springer}
}

\newpage

\vfill

\end{document}